\documentclass[10pt,twocolumn,letterpaper]{article}
\usepackage{cvpr}              

\usepackage[export]{adjustbox}
\definecolor{cvprblue}{rgb}{0.21,0.49,0.74}
\usepackage[pagebackref,breaklinks,colorlinks,allcolors=cvprblue]{hyperref}

\usepackage[ruled]{algorithm2e}
\usepackage[table]{xcolor}
\usepackage{multirow}
\usepackage{graphicx}
\usepackage{booktabs}
\usepackage{amsmath}
\usepackage{bm}
\usepackage{mathtools}
\usepackage{pifont}
\usepackage{xspace}

\definecolor{theyellow}{RGB}{255,240,193}
\colorlet{myyellow}{theyellow!88}
\colorlet{lightyellow}{theyellow!56}
\definecolor{DarkBlue}{RGB}{0,0,128}
\definecolor{LightBlue}{RGB}{64,101,149}
\definecolor{mygray}{RGB}{230,235,235}
\usepackage{threeparttable}
\newcolumntype{I}{!{\vrule width 0.8pt}}
\makeatletter
\newcommand{\thickhline}{%
    \noalign {\ifnum 0=`}\fi \hrule height 0.5pt
    \futurelet \reserved@a \@xhline
}
\makeatother
\definecolor{RedOrange}{rgb}{1.0, 0.27, 0.0}

\newcommand{\myred}[1]{$_{\color{RedOrange}\uparrow #1}$}
\newcommand{\pub}[1]{{\color{gray}{\scriptsize{[{#1}]}}}}
\definecolor{DeepRed}{rgb}{0.8, 0, 0}

\newcommand{\my}{$F^2$DC\xspace}
\newcommand{\rf}{$f_{i}^{\scalebox{0.66}{$\bm{+}$}}$\xspace}
\newcommand{\equrf}{f_{i}^{\scalebox{0.66}{$\bm{+}$}}\xspace}
\newcommand{\nrf}{$f_{i}^{\scalebox{0.66}{$\bm{-}$}}$\xspace}
\newcommand{\equnrf}{f_{i}^{\scalebox{0.66}{$\bm{-}$}}\xspace}
\newcommand{\corf}{$f_{i}^{\scalebox{0.88}{$\star$}}$\xspace}
\newcommand{\equcorf}{f_{i}^{\scalebox{0.88}{$\star$}}\xspace}

\newcommand{\equnewf}{\widetilde{f_{i}}\xspace}

\def\paperID{9515}
\def\confName{CVPR}
\def\confYear{2026}

\title{Domain-Skewed Federated Learning with Feature Decoupling and Calibration}

\author{Huan Wang\textsuperscript{1,2} \quad Jun Shen\textsuperscript{1}$^*$ \quad Jun Yan\textsuperscript{1} \quad Guansong Pang\textsuperscript{2}\thanks{Corresponding Authors: Jun Shen, Guansong Pang}\\
\textsuperscript{1}School of Computing and Information Technology, University of Wollongong, Australia \\
\textsuperscript{2}School of Computing and Information Systems, Singapore Management University, Singapore \\
{\tt\small hw226@uowmail.edu.au, jshen@uow.edu.au, jyan@uow.edu.au, gspang@smu.edu.sg}
}

\begin{document}
\maketitle
\begin{abstract}
Federated learning (FL) allows distributed clients to collaboratively train a global model in a privacy-preserving manner. 
However, one major challenge is domain skew, where clients' data originating from diverse domains may hinder the aggregated global model from learning a consistent representation space, resulting in poor generalizable ability in multiple domains. 
In this paper, we argue that the domain skew is reflected in the domain-specific biased features of each client, causing the local model's representations to collapse into a narrow low-dimensional subspace. 
We then propose \textbf{F}ederated \textbf{F}eature \textbf{D}ecoupling and \textbf{C}alibration (\my), which liberates valuable class-relevant information by calibrating the domain-specific biased features, enabling more consistent representations across domains. 
A novel component, Domain Feature Decoupler (DFD), is first introduced in \my to determine the robustness of each feature unit, thereby separating the local features into domain-robust features and domain-related features. 
A Domain Feature Corrector (DFC) is further proposed to calibrate these domain-related features by explicitly linking discriminative signals, capturing additional class-relevant clues that complement the domain-robust features. 
Finally, a domain-aware aggregation of the local models is performed to promote consensus among clients. 
Empirical results on three popular multi-domain datasets demonstrate the effectiveness of the proposed \my and the contributions of its two modules. 
Code is available at \url{https://github.com/mala-lab/F2DC}.
\end{abstract}

\section{Introduction}
Federated Learning (FL) \cite{mcmahan2017communication,li2020federated} is emerging as a privacy-preserving technique for collaboratively training a shared global model on multiple clients without leaking privacy. The FL approaches are embodied by key cornerstone algorithms such as FedAvg \cite{mcmahan2017communication}, where each client executes stochastic gradient descent iterations to minimize the local empirical objective, and all participating clients' local models are then sent to the server to perform global aggregation of parameters. An inherent challenge in the FL paradigm is the potential discrepancies of local data distributions among clients, referred to as \textit{data heterogeneity} \cite{zhu2021federated,li2022federated,kairouz2021advances,li2020federated}. 
In particular, clients have drastically different data preferences collected from distinct sources, resulting in non-independent and identically distributed (non-iid) data distribution \cite{zhu2021federated,li2022federated}. 
Such data heterogeneity has caused the global model to perform poorly \cite{li2020fedprox,Li2020On}.

\begin{figure}[t]
    \centering
    \includegraphics[width=0.9\linewidth]{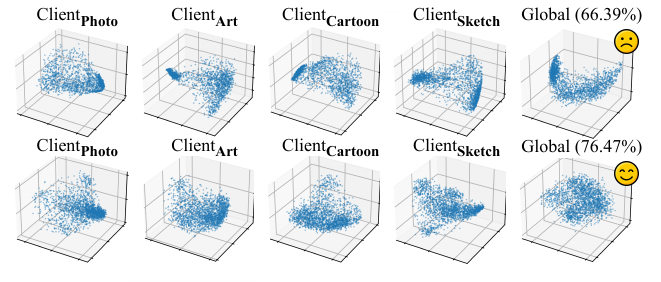}
    \caption{Visualization of the singular values of the feature covariance matrix on the PACS \cite{li2017deeper} dataset from Vanilla FL \cite{mcmahan2017communication} (\textbf{Top}) and our method (\textbf{Bottom}). Under domain-skewed FL, the Vanilla method suffers from severe dimensional collapse, \ie, representations are largely biased and reside in a domain-specific lower-dimensional subspace. Our method effectively mitigates this collapse, as reflected in more uniformly distributed singular values and significantly fewer values tending to zero.}
\label{fig:fig1}
\end{figure}

\begin{figure}[t]
    \centering
    \includegraphics[width=0.88\linewidth]{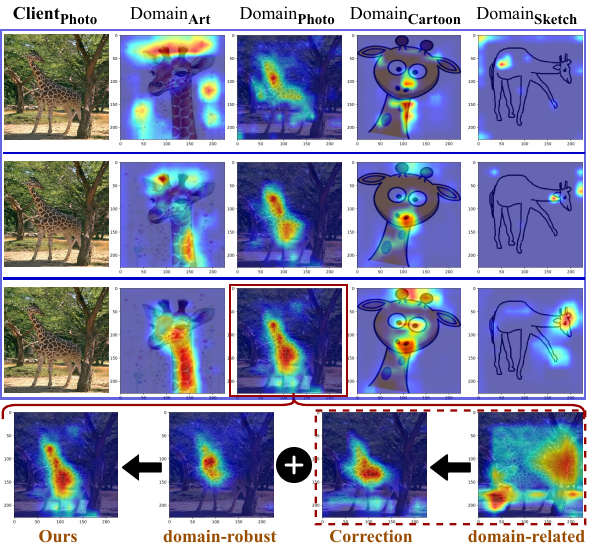}
    \caption{Grad-CAM \cite{selvaraju2017grad} visualization of the features obtained by three methods—Vanilla method \cite{mcmahan2017communication} (\textbf{Top}), domain-skewed method FDSE \cite{wang2025federated} (\textbf{Middle}), and \my (\textbf{Bottom})—for samples from different domains on a randomly selected $\text{Client}_{Photo}$ belonging to the photo domain. 
    For our method \my, we further provide a qualitative example about how its two modules work: local features are first decoupled via DFD into domain-robust features and domain-related features. Then, the domain-related features (which entangle valuable class-relevant information with domain biases) are calibrated via DFC, resulting in the Correction features. By incorporating the domain-robust and the corrected features, \my captures more holistic class-relevant clues, enabling the local model to yield more consistent decisions for samples with the same semantics across different domains.}
    \label{fig:fig2}
\end{figure}

To alleviate the data heterogeneity problem, subsequent studies focus on stabilizing local training \cite{li2020fedprox,karimireddy2020scaffold,wang2025feddifrc} or regulating global aggregation \cite{ye2023feddisco,wang2024aggregation,wang2023fedcda} to reduce the variance in the FL optimization. 
Notably, these methods primarily consider a \textit{label skew} issue, \ie, data class distributions are different among clients while all data originate from the same domain. 
However, for many real-world scenarios (\eg, driving data collected from different vehicles under diverse weather conditions), the data heterogeneity is predominated by \textit{\textbf{domain skew}}, where the local data among different clients are drawn from varying domains but have a similar class distribution. 
This often leads to severe \textit{dimensional collapse} in local model training due to their exclusive fitting to the features of individual domains in each client, and consequently, the representations in each client mainly reside in a narrow low-dimensional subspace, ignoring relevant features in the other subspaces. 
This issue manifests as small (near-zero) singular values of the feature covariance matrix in each client, as illustrated in Fig.~\ref{fig:fig1}-\textbf{Top}. 
Aggregating from such domain-biased local models can result in a great difficulty to obtain consensus updates in the global model, rendering it ineffective in generalizing to data from different domains, as exemplified by the large discrepancy in the attended features across the domains in Fig.~\ref{fig:fig2}-\textbf{Top}. 
This motivates the question: \textit{Can we promote local training away from client-specific domain preferences under the domain-skewed FL scenario?}

To address this challenge, recent advanced FL methods \cite{wang2025federated,fu2025federated} aim to eliminate and deactivate domain-specific biases from the local parameters to improve performance and facilitate consensus updates. 
However, these `elimination-based' methods risk discarding additional class-relevant information that is entangled with these domain-specific features, \eg, FDSE \cite{wang2025federated} misses the antlers and head of the giraffe in the cartoon and sketch domains in Fig.~\ref{fig:fig2}-\textbf{Middle}. 
This motivates our \textbf{core question}: \textit{Can we correct domain-specific biases to capture additional class-relevant features, thus promoting a more domain-generalizable global model?}

To overcome this domain skew issue, we propose \textbf{\textit{F}$\mathbf{^2}$DC}, namely \textbf{F}ederated \textbf{F}eature \textbf{D}ecoupling and \textbf{C}alibration, aiming to correct domain-specific biased features to improve the generalizable ability of local training. 
\my consists of two novel modules, including Domain Feature Decoupler (DFD) and Domain Feature Corrector (DFC). 
DFD is optimized to decouple the local features into \textit{domain-robust features} that focus on discriminative cross-domain class-relevant information, and \textit{domain-related features} that are responsible for the client-specific domain context. This prevents \my from overfitting the domain-biased features, as shown in Fig.~\ref{fig:fig1}-\textbf{Bottom}. 
Further, the domain-related features inherently entangle domain biases (\eg, the stroke style of a sketch domain) with class-relevant information (\eg, the `object outline' formed by the sketch). 
To capture these additional class-relevant clues that can complement the domain-robust features, DFC is devised to calibrate these domain-related features by explicitly linking correct semantic signals. 
Finally, we perform domain-aware aggregation of the local models to incorporate the domain discrepancy into the global model, thereby further promoting consensus among clients. 
As shown in Fig.~\ref{fig:fig2}-\textbf{Bottom}, \my can more consistently obtain discriminative features across domains compared to \cite{mcmahan2017communication,wang2025federated}; to zoom in the photo domain example, DFD in \my accurately decouples the domain-robust features (the upper giraffe body) from the domain-related features, and it further captures the additional class-relevant clues (\eg, correction features recapture the giraffe's waist) from the noisy domain-related features via DFC. 
The main contributions are listed below:
\begin{itemize}
    \item Rather than eliminating domain-specific biases as recent domain-skewed FL methods do, we argue that these biases are inherently entangled with valuable class-relevant information, once properly calibrated, can be leveraged to produce more consistent decisions across domains.
    \item We propose \my, a novel solution for domain-skewed FL. It improves cross-domain performance by decoupling and correcting the local features through its two novel modules DFD and DFC, while performing domain-aware aggregation to promote more consensus among clients.
    \item Extensive experiments conducted on three multi-domain datasets demonstrate that \my consistently outperforms state-of-the-art methods. Comprehensive ablative studies also confirm the contributions of its two modules.
\end{itemize}

\section{Related Work}
\paragraph{Heterogeneous Federated Learning.} Federated learning has been an emerging topic that collaboratively trains a shared global model without sharing private data \cite{mcmahan2017communication,li2020fedprox,li2022federated}. However, due to the data heterogeneity, this paradigm suffers from unstable, slow convergence \cite{huang2024federated,zhu2021federated,li2022federated,gao2022feddc,Li2020On}. 
To tackle this, previous studies either regularize local training at the client-side \cite{huang2022learn,huang2023rethinking,wang2025fedsc,wang2025fedskc,xu2025federated} or adjust global aggregation at the server-side \cite{zhang2023fedala,ye2023feddisco,wang2023fedcda,wang2024aggregation,huang2025fedbg}. 
However, most of these methods neglect the domain skew, leading to compromised performance on multiple domains \cite{chen2024fair,huang2023rethinking}.

\paragraph{Domain-Skewed Federated Learning.} Domain-skewed data readily weakens the generalization of the global model, which has motivated the development of domain-skewed FL \cite{sun2021partialfed,huang2023rethinking,wang2025federated,zhou2025fedsa,fu2025beyond,chen2024fair,zhang2024fedgmkd}. 
To mitigate the domain skew, there are mainly two directions: enhancing the uniformity between clients \cite{huang2023rethinking,chen2024fair,jiang2022harmofl,zhou2025fedsa} and personalizing the global model \cite{t2020personalized,li2025personalized,li2024personalized}. 
FedSA \cite{zhou2025fedsa} builds anchor-based regularization to ensure consistent boundaries across domains. 
FedHEAL \cite{chen2024fair} leverages selective parameter local updating and a fair aggregation objective to facilitate fairness and performance. 
FDSE \cite{wang2025federated} decouples each model layer into a domain-agnostic extractor and a domain-specific eraser, achieving global consensus and local personalization. 
However, these methods aim to discard or deactivate the domain-biased parameter updates of the local training, ignoring useful clues that may exist in the domain context. 
In contrast, we claim that, with proper correction, the domain-related features of each client can recapture additional useful guidance to promote cross-domain generalizability.

\paragraph{Disentangled Representation Learning.} Disentangled representation learning (DRL) aims to learn a mechanism to identify and disentangle latent factors hidden in the data at the feature-level \cite{wang2024disentangled,zhang2019gait,fumero2023leveraging}. Previous works are focused on separating general and individual features for downstream tasks, such as gait recognition \cite{li2020gait,zhang2019gait}, recommendation \cite{ma2019learning}, graph analysis~\cite{mo2023disentangled,zhang2023dyted}, etc. 
There are also some works that incorporate disentangled learning into federated learning \cite{bercea2022federated,bai2024diprompt,uddin2022federated,yan2023personalization,chen2024disentanglement} to assist local training in FL to achieve higher performance. 
However, these methods focus on single-domain performance. Under a domain skew FL scenario, it is essential to consider generalization across different domains. Our work illustrates how to leverage the disentangled learning (\ie, local feature decoupling and correction) to tackle the domain skew challenge in FL.

\section{Problem Statement} \label{sec3}
Following~\cite{huang2023rethinking,chen2024fair}, we focus on the standard domain-skewed FL scenario for $K$ participating clients associated with data partitions $\{\mathcal{D}_{1}, \ldots, \mathcal{D}_{k}, \ldots, \mathcal{D}_{K}\}$, where each client's data samples $\mathcal{D}_{k}=\{x_{i}, y_{i}\}|_{i=1}^{n_{k}}, y_{i} \in \{1, \ldots, \mathbb{C}\}$ only belonging to a single domain, $n_{k}$ denotes the local data scale for client $k$, and $\mathbb{C}$ denotes the number of categories. 
Let $n_{k}^{c}$ be the number of samples with class $c$ on client $k$, we further define $\mathcal{D}_{k}^{c}=\{(x_{i},y_{i}) \in \mathcal{D}_{k}|y_{i}=c\}$ as the set of samples with class $c$ on client $k$, where $n_{k}^{c} = |\mathcal{D}_{k}^{c}|$. The number of samples in class $c$ of all clients is $n^{c} = \sum_{k=1}^{K} n_{k}^{c}$. 
Formally, the optimization objective of the domain-skewed FL is:
\begin{equation} \label{eq:eq1} 
\setlength\abovedisplayskip{3pt} \setlength\belowdisplayskip{3pt} 
    w^{*} = \arg\min_{w} \mathcal{L}(w) = \sum_{k=1}^{K} \frac{n_{k}}{N} \mathcal{L}_{k}(w;\mathcal{D}_{k}),
\end{equation}
where $w$ is the parameters of the global model, $\mathcal{L}_{k}$ denotes the empirical objective on the client $k$, and $N=\sum_{k=1}^{K} n_{k}$ denotes the total number of samples among all clients. 
The optimization objective in Eq.~\eqref{eq:eq1} is to learn a generalizable global model $w^{*}$ to present favorable performance on multiple domains during the FL process. 
In the FL training setup, for each client $k$, we consider the local model with parameters $w_{k}=\{\Phi_{k}, \theta_{k}\}$. It has two modules: a feature extractor $r(\Phi_{k}; \cdot): x \rightarrow \mathbb{R}^{d}$, that encodes each sample $x$ to a $d$-dim embedding $z=r(\Phi_{k}; x)$; a classifier $h(\theta_{k}; \cdot): \mathbb{R}^{d} \rightarrow \mathbb{R}^{\mathbb{C}}$, that maps the embedding $z$ into a $\mathbb{C}$-dim logit $\ell=h(\theta_{k}; z)$.

\paragraph{Domain Skew.} In heterogeneous FL scenario, the feature distributions $\mathbb{P}(x|y)$ differ across clients even when $\mathbb{P}(y)$ is consistent, leading to the domain skew problem~\cite{huang2023rethinking}:
\begin{equation} \label{eq:eq2} 
\setlength\abovedisplayskip{3pt} \setlength\belowdisplayskip{3pt} 
    \mathbb{P}_{k_{1}}(x|y) \neq \mathbb{P}_{k_{2}}(x|y), \;\; \text{s.t.} \;\;\; \mathbb{P}_{k_{1}}(y)=\mathbb{P}_{k_{2}}(y).
\end{equation}

\section{Methodology}
\begin{figure*}[t]
    \centering
    \includegraphics[width=0.90\linewidth]{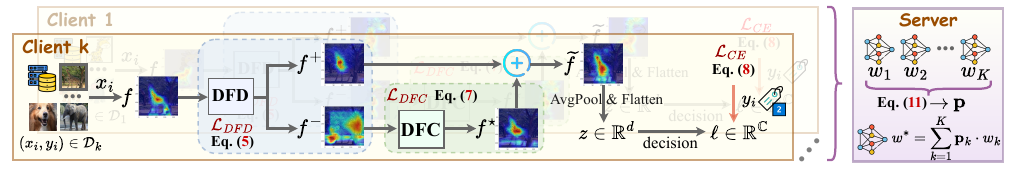}
    \caption{\textbf{Overview} of the proposed \my approach. During client-side local training, we initially separate local features into domain-robust features and domain-related features via a Domain Feature Decoupler (DFD) (Sec.~\ref{sec41}), and then rectify the domain-related features via a Domain Feature Corrector (DFC) (Sec.~\ref{sec42}). During server-side global aggregation, we perform domain-aware aggregation (Sec.~\ref{sec43}) of the global model by incorporating each client's domain discrepancy $\mathbf{p}_{k}$ to promote more consensus among local clients.}
    \label{fig:fig3}
\end{figure*}

To promote cross-domain generalization ability in domain-skewed FL, we propose \my. As illustrated in Fig.~\ref{fig:fig3}, \my consists of two main modules, Domain Feature Decoupler (DFD) and Domain Feature Corrector (DFC). First, DFD determines the robustness of each feature unit, thereby separating the local features into domain-robust features and domain-related features (Sec.~\ref{sec41}). 
Then, DFC calibrates the domain-related features, enabling the local model to capture additional useful clues for improving decisions across domains (Sec.~\ref{sec42}). Finally, we perform global aggregation in a domain-aware manner to encourage consensus among local clients (Sec.~\ref{sec43}).

\subsection{Domain Feature Decoupler (DFD)} \label{sec41}
Our key insight lies in calibrating biased domain-related features during the local training. 
To this end, our initial step is to isolate these domain-related features from the local features, which capture the client-specific domain context for subsequent calibration. 
With the above definitions in Sec.~\ref{sec3}, for the client $k$, we further refine the feature extractor $r(\Phi_{k}; \cdot)$ as backbone layers $r_{k}^{\mathtt{B}}$ and flatten layers $r_{k}^{\mathtt{F}}$: taking ResNet-10 \cite{he2016deep} as an example, $r_{k}^{\mathtt{B}}$ includes $4$ backbone layers $L_{1} \sim L_{4}$, $r_{k}^{\mathtt{F}}$ consists of a pooling layer and a flattening operation. 
Given an instance $(x_{i}, y_{i}) \in \mathcal{D}_{k}$, we first calculate the feature map $f_{i}=r_{k}^{\mathtt{B}}(x_{i}) \in \mathbb{R}^{C \times H \times W}$ (\eg, for ResNet-10 \cite{he2016deep} on PACS \cite{li2017deeper}, $f_{i} \in \mathbb{R}^{512 \times 16 \times 16}$).

To determine which feature unit of $f_{i}$ is domain-robust or domain-related, we introduce a Domain Feature Decoupler (DFD) $\mathcal{A}_{D}$ to construct an attribution map $\mathcal{S}_{i}=\mathcal{A}_{D}(f_{i}) \in \mathbb{R}^{C \times H \times W}$, which quantifies cross-domain robustness score of each feature unit in the $f_{i}$. 
We implement the decoupler $\mathcal{A}_{D}$ as a two-layer CNN architecture with batch normalization and ReLU activation. 
To decompose $f_{i}$ in a complementary manner, we further generate a binary mask matrix $\mathcal{M}_{i}=\{0, 1\}^{C \times H \times W}$ based on $\mathcal{S}_{i}$. 
The direct strategy is to binarize each element in $\mathcal{S}_{i}$ into $0$ or $1$, but such hard discretization would lead to non-differentiability in training \cite{cai2020rethinking}. 
To avoid this problem, we leverage a Gumbel concrete distribution \cite{louizos2018learning} to generate $\mathcal{M}_{i}$, defined as:
\begin{equation} \label{eq:eq3}
\setlength\abovedisplayskip{3pt} \setlength\belowdisplayskip{3pt}
    \mathcal{M}_{i} = \frac{\mathtt{e}^{\frac{1}{\sigma} (\log(\epsilon(\mathcal{S}_{i}))+\mathbf{g}_{a})}}{
    \mathtt{e}^{\frac{1}{\sigma} (\log(\epsilon(\mathcal{S}_{i}))+\mathbf{g}_{a})} + 
    \mathtt{e}^{\frac{1}{\sigma} (\log(1-\epsilon(\mathcal{S}_{i}))+\mathbf{g}_{b})}},
\end{equation}
where $\mathbf{g}$ is a Gumbel noise $\mathbf{g}=\log(u)-\log(1-u)$ and $u \sim \mathcal{U}(0,\mathbf{I})$, $\sigma$ is used to adjust the smoothness of $\mathbf{g}_{a}$ and $\mathbf{g}_{b}$ ($\sigma \rightarrow 0$, $\mathcal{M}_{i}$ is approximately a hard discretization), $\epsilon$ is a Sigmoid function, and $\mathtt{e}$ denotes an Exponential function.

On the basis of Eq.~\eqref{eq:eq3}, a pseudo-binary mask matrix $\mathcal{M}_{i}$ is generated based on the Gumbel score of $\mathcal{S}_{i}$, where the element's value of $\mathcal{M}_{i}$ close to $1$ indicates a domain-robust feature unit and close to $0$ as a domain-related feature unit. 
We then decouple the local feature map $f_{i}$ via $\mathcal{M}_{i}$ as:
\begin{equation} \label{eq:eq4} 
\setlength\abovedisplayskip{3pt} \setlength\belowdisplayskip{3pt} 
    \equrf = \mathcal{M}_{i} \odot f_{i}, \;\; \equnrf = (1-\mathcal{M}_{i}) \odot f_{i},
\end{equation}
where $\odot$ represents an element-wise product operator, the \rf and \nrf are domain-robust and domain-related features.

On the one hand, we expect to ensure decent separation between \rf and \nrf to provide class-relevant information and biased domain context. On the other hand, we emphasize maintaining discriminative ability. Consequently, we encourage the learning of $\mathcal{A}_{D}$ by:
\begin{equation} \label{eq:eq5} 
\setlength\abovedisplayskip{3pt} \setlength\belowdisplayskip{3pt} 
\begin{aligned}
    \mathcal{L}_{DFD} = & \underbrace{\log(\exp(\mathbf{s}(l_{i}^{\scalebox{0.66}{$\bm{+}$}}, l_{i}^{\scalebox{0.66}{$\bm{-}$}})/\tau))}_{\textbf{Separability}} - \\
    & \underbrace{( y_{i} \cdot \log(\delta(\mathbf{m}(l_{i}^{\scalebox{0.66}{$\bm{+}$}}))) + \widehat{y_{i}} \cdot \log(\delta(\mathbf{m}(l_{i}^{\scalebox{0.66}{$\bm{-}$}}))) )}_{\textbf{Discriminability}}, \\
    \multicolumn{2}{c}{$\text{s.t.} \;\;\; l_{i}^{\scalebox{0.66}{$\bm{+}$}}=r_{k}^{\mathtt{F}}(\equrf) \in \mathbb{R}^{d}, \, l_{i}^{\scalebox{0.66}{$\bm{-}$}}=r_{k}^{\mathtt{F}}(\equnrf) \in \mathbb{R}^{d},$} \\
    \multicolumn{2}{c}{$\text{s.t.} \;\;\; \mathbf{s}(l_{i}^{\scalebox{0.66}{$\bm{+}$}}, l_{i}^{\scalebox{0.66}{$\bm{-}$}}) = 
    (l_{i}^{\scalebox{0.66}{$\bm{+}$}} \cdot l_{i}^{\scalebox{0.66}{$\bm{-}$}}) \bm{/} 
    (\|l_{i}^{\scalebox{0.66}{$\bm{+}$}}\|_{2} \times \|l_{i}^{\scalebox{0.66}{$\bm{-}$}}\|_{2}),$}
\end{aligned}
\end{equation}
where $r_{k}^{\mathtt{F}}$ consists of a pooling layer and a flattening operation as $\mathbb{R}^{C \times H \times W} \rightarrow \mathbb{R}^{d}$, $\mathbf{s}$ denotes calculating cosine similarity, $\mathbf{m}$ is a trainable single-layer MLP architecture as $\mathbb{R}^{d} \rightarrow \mathbb{R}^{\mathbb{C}}$ to predict the logits of \rf and \nrf, $\widehat{y_{i}}$ represents the wrong label with the highest confidence except the ground truth $y_{i}$, $\tau$ is used to control the concentration strength of representations \cite{wang2021understanding} to help suitably separate the \rf and \nrf, and $\delta$ represents a Softmax function.

By minimizing Eq.~\eqref{eq:eq5}, we optimize $\mathcal{A}_{D}$ to assign higher elements to \rf via $\mathcal{M}_{i}$ in Eq.~\eqref{eq:eq3}, which encourages $\mathbf{m}$ to make correct decisions. 
Meanwhile, $\mathcal{A}_{D}$ reduces the attention of domain-related features \nrf to the semantic object, shifting the focus toward the client-specific domain context that may cause wrong decisions. 
Notably, this domain context can contain some inherently entangled class-relevant information and noisy domain biases, on which $\mathcal{A}_{D}$ cannot perfectly isolate all discriminative signals into \rf, \ie, $\mathcal{A}_{D}$ inevitably relegates those `mixed' features into \nrf. 
Therefore, the resulting \nrf is often a complex mixture of domain artifacts and valuable class-relevant clues. 
This point is empirically verified in two directions: 1) enforcing an overly aggressive separation, \eg, via using a small $\tau$ ($\tau=0.02$) to learn all discriminative signals into \rf, does not yield more optimal performance, as shown in Fig. \ref{fig:fig6}; and 2) the resulting \corf (calibrated from \nrf) yields significant performance improvement, proving that \nrf is rich with entangled class-relevant clues, as indicated in Table \ref{tab:tab7}.

\subsection{Domain Feature Corrector (DFC)} \label{sec42}
We then aim to calibrate the domain-related features \nrf to capture additional discriminative clues that may complement \rf, thereby promoting the local model to yield more consistent decisions for samples with the same class across domains. 
To this end, we design a Domain Feature Corrector (DFC) $\mathcal{A}_{C}$ that takes the domain-related features \nrf as input and produces their rectified features \corf. We implement the corrector $\mathcal{A}_{C}$ as the same two-layer CNN architecture as the decoupler $\mathcal{A}_{D}$. In particular, $\mathcal{A}_{C}$ learns the residual term \cite{he2016deep} for more flexible training:
\begin{equation} \label{eq:eq6} 
\setlength\abovedisplayskip{3pt} \setlength\belowdisplayskip{3pt} 
    \equcorf = \equnrf + (1-\mathcal{M}_{i}) \odot \mathcal{A}_{C}(\equnrf),
\end{equation}
where $\mathcal{M}_{i}$ is the pseudo-binary mask matrix in Eq.~\eqref{eq:eq3}, and $\odot$ represents an element-wise product operator.

The intended goal of our corrector is to enable rectified features \corf to provide useful clues that steer the local model toward more correct decisions. 
Therefore, we optimize the corrector $\mathcal{A}_{C}$ by explicitly injecting correct discriminative semantic signals, which is formulated as:
\begin{equation} \label{eq:eq7} 
\setlength\abovedisplayskip{3pt} \setlength\belowdisplayskip{3pt} 
    \mathcal{L}_{DFC} = - y_{i} \cdot \log(\delta(\mathbf{m}(l_{i}^{\scalebox{0.88}{$\star$}}))), \;\; 
    \text{s.t.} \;\;\; l_{i}^{\scalebox{0.88}{$\star$}}=r_{k}^{\mathtt{F}}(\equcorf),
\end{equation}
where $r_{k}^{\mathtt{F}}$ consists of a pooling layer and a flattening operation to encode \corf as $l_{i}^{\scalebox{0.88}{$\star$}} \in \mathbb{R}^{d}$, $\delta$ is a Softmax function, and $\mathbf{m}$ is the same as the single-layer MLP in Eq.~\eqref{eq:eq5}. 
By minimizing Eq.~\eqref{eq:eq7}, we promote the corrector $\mathcal{A}_{C}$ to regulate the domain-related features \nrf to the rectified features \corf, thereby capturing additional class-relevant information. 
We then combine the domain-robust features \rf and the rectified features \corf to obtain the final feature map $\equnewf = (\equrf + \equcorf) \in $ $\mathbb{R}^{C \times H \times W}$, which is further passed to the subsequent layers of the local model.

Besides, the standard Cross-Entropy \cite{de2005tutorial} loss is used on top of the logits $\ell_{i}$ with $y_{i}$ to optimize the local model:
\begin{equation} \label{eq:eq8} 
\setlength\abovedisplayskip{3pt} \setlength\belowdisplayskip{3pt} 
    \mathcal{L}_{CE} = - \mathbf{1}_{y_{i}} \log(\delta(\ell_{i})), \;\; 
    \text{s.t.} \;\;\; \ell_{i}=h(\theta_{k}; r_{k}^{\mathtt{F}}(\equnewf)),
\end{equation}
where $\delta$ represents a Softmax function, and $h(\theta_{k}; \cdot)$ is the classifier of the local model. Overall, we define the following objective for the local training of each client $k$ as:
\begin{equation} \label{eq:eq9} 
\setlength\abovedisplayskip{3pt} \setlength\belowdisplayskip{3pt} 
    \mathcal{L} = \mathcal{L}_{CE} + \frac{1}{|L|} \sum_{j=1}^{|L|} (\lambda_{1} \cdot \mathcal{L}_{DFD}^{L_{j}} + \lambda_{2} \cdot \mathcal{L}_{DFC}^{L_{j}}),
\end{equation}
where $\mathcal{L}_{DFD}^{L_{j}}$ and $\mathcal{L}_{DFC}^{L_{j}}$ mean the losses $\mathcal{L}_{DFD}$ and $\mathcal{L}_{DFC}$ calculated after the $L_{j}$ backbone layer of the local model's feature extractor, respectively. 
By default, we only calculate the losses $\mathcal{L}_{DFD}$ and $\mathcal{L}_{DFC}$ after the last backbone layer of the feature extractor ($|L|=1$) for each client. 
$\lambda_{1}$ and $\lambda_{2}$ are hyper-parameters used to balance these loss terms.

\subsection{Domain-Aware Aggregation} \label{sec43}
During the global aggregation phase, under domain-skewed FL scenario, the solution of naively performing average aggregation \cite{mcmahan2017communication} is biased and neglects the domain diversity, resulting in restricted model performance. 
To mitigate this issue, we consider incorporating the domain discrepancy of each client into the global model's aggregation process in a domain-aware manner. 
We first establish an ideal reference: a uniform global domain distribution, as it naturally promotes fairness across all domains for the same class. 
Then, we measure the distance between each local domain distribution and this uniform global domain distribution, treating it as a domain discrepancy.

Specifically, we define this uniform global domain distribution as a fixed vector $\mathcal{G}=[\frac{1}{Q}, \ldots, \frac{1}{Q}] \in \mathbb{R}^{\mathbb{C}}$, where $Q$ denotes the number of domains and the value of each element in $\mathcal{G}$ is $1/Q$ (\textit{e.g.}, for PACS \cite{li2017deeper}, $Q=4$). 
For client $k$, the local domain distribution $\mathcal{B}_{k}=[\frac{n_{k}^{1}}{n^{1}}, \ldots, \frac{n_{k}^{\mathbb{C}}}{n^{\mathbb{C}}}] \in \mathbb{R}^{\mathbb{C}}$ is also abstracted as a vector, where $n_{k}^{c} \approx n_{k} / \mathbb{C}$ and $n^{c} \approx N / \mathbb{C}$ (under the domain skew, the label distribution is consistent). 
Therefore, we reformulate the local domain distribution of the client $k$ as $\mathcal{B}_{k}=[\frac{n_{k}}{N}, \ldots, \frac{n_{k}}{N}] \in \mathbb{R}^{\mathbb{C}}$. 
Then, we calculate the domain discrepancy $\mathbf{d}_{k}$ of the client $k$ as follows:
\begin{equation} \label{eq:eq10} 
\setlength\abovedisplayskip{3pt} \setlength\belowdisplayskip{3pt} 
    \mathbf{d}_{k} = \sqrt{\frac{1}{2} \sum_{c=1}^{\mathbb{C}} \left( \mathcal{B}_{k}^{c} - \mathcal{G}^{c} \right)^{2}}, \;\;\; \mathcal{B}_{k}^{c}=\frac{n_{k}}{N}, \;\; \mathcal{G}^{c}=\frac{1}{Q},
\end{equation}
where $Q$ means the number of domains and $N=\sum_{k=1}^{K} n_{k}$. Based on Eq.~\eqref{eq:eq10}, we obtain the domain discrepancy $\mathbf{d}_{k}$ for each client $k$. We then enforce that the client $k$ with a larger dataset size $n_{k}$ and smaller domain discrepancy $\mathbf{d}_{k}$ receives a higher aggregation weight $\mathbf{p}_{k}$:
\begin{equation} \label{eq:eq11} 
\setlength\abovedisplayskip{3pt} \setlength\belowdisplayskip{3pt} 
    \mathbf{p}_{k} = \frac{\epsilon(\alpha \cdot \frac{n_{k}}{N} - \beta \cdot \mathbf{d}_{k})}{
    \sum_{j=1}^{K} \epsilon(\alpha \cdot \frac{n_{j}}{N} - \beta \cdot \mathbf{d}_{j})},
\end{equation}
where $\epsilon$ denotes Sigmoid function, $\alpha$ and $\beta$ are used to balance the $n_{k}$ and $\mathbf{d}_{k}$. 
According to Eq.~\eqref{eq:eq11}, even if the client dataset size is large (\eg, the dominant domain), we can determine a more distinguishing weight for each client rather than a simple average. Finally, we obtain the global model $w^{*} = \sum_{k=1}^{K} \mathbf{p}_{k} \cdot w_{k}$ in a domain-aware manner.

The global model is then broadcast to each participating client as the initialization for the next round's local updating. 
Note that each client decoupler $\mathcal{A}_{D}$, corrector $\mathcal{A}_{C}$, and $\mathbf{m}$ are kept locally, without participating in aggregation.

\subsection{Summary and Discussion} \label{secsum}
In each communication round, the local model is trained on its private dataset by optimizing $\mathcal{L}$ in Eq.~\eqref{eq:eq9}, then the server collects these local weights to obtain the global model in a domain-aware aggregation manner, which is subsequently distributed to each participant as the initialization in the next communication round. 
The overview of \my is shown in Fig.~\ref{fig:fig3}, and a detailed algorithm is presented in Algorithm~\ref{alg:f2dc}.

Besides, the modularity of our proposed \my suggests its broad scope of application. 
On the one hand, our decoupler $\mathcal{A}_{D}$ and corrector $\mathcal{A}_{C}$ can be easily incorporated with existing FL approaches to promote more robust local decisions. On the other hand, integrating the domain discrepancy into the global aggregation helps improve consensus among clients for the domain-skewed FL scenario. 
Empirically, enhancing FedAvg \cite{mcmahan2017communication} with only our domain feature decoupler and corrector yields $8.94\%$ performance gain on PACS \cite{li2017deeper}, please refer to the details in Table~\ref{tab:tab3}.

\begin{algorithm}[t]
\caption{\small{Pseudo-code Flow of \my}}
\label{alg:f2dc}

\SetAlgoLined
\SetNoFillComment
\SetArgSty{textnormal}

\small{\KwIn{Communication Rounds $R$, Local Epochs $E$, Clients $K$, $k$-th client's data $\mathcal{D}_{k}$ and model $w_{k}$}}
\small{\KwOut{The final FL global model $w_{R}^{*}$}}


\For {each communication round $r=1, 2, ..., R$}{
    $\mathcal{K}_{r} \Leftarrow \{1, ..., K\}$ {\footnotesize{\color{DarkBlue}{\tcp{randomly select clients}}}}
    
    \For {each client $k \in \mathcal{K}_{r}$ \textbf{in parallel}}{
        $w_{k} \leftarrow \;$ \KwSty{LocalUpdating}($w_{r}^{*}$)
    }

    {\footnotesize{\color{DarkBlue}{\tcc{Domain-Aware Aggregation (Sec.\ref{sec43})}}}}
    
    $\mathbf{d}_{k} \leftarrow (\mathcal{B}_{k}, \mathcal{G})$ in Eq.~\eqref{eq:eq10} {\footnotesize{\color{DarkBlue}{\tcp{discrepancy}}}}

    $\mathbf{p}_{k} \leftarrow (\mathbf{d}_{k}, n_{k})$ in Eq.~\eqref{eq:eq11} {\footnotesize{\color{DarkBlue}{\tcp{determine weight}}}}

    $w_{r+1}^{*} \leftarrow \sum_{k=1}^{|\mathcal{K}_{r}|} \mathbf{p}_{k} \cdot w_{k}$ {\footnotesize{\color{DarkBlue}{\tcp{global model}}}}
}


\KwSty{LocalUpdating}($w_{r}^{*}$):

$w_{k} \leftarrow w_{r}^{*} + \{\mathcal{A}_{D}, \mathcal{A}_{C}, \mathbf{m}\}$ {\footnotesize{\color{DarkBlue}{\tcp{distribute}}}}

\For {each local epoch $e=1, 2, ..., E$}{
    \For {each batch $b \in $ private data $\mathcal{D}_{k}$}{
    
        {\footnotesize{\color{DarkBlue}{\tcc{DFD Module (Sec.\ref{sec41})}}}}

        $\mathcal{M}_{i} \leftarrow (\mathcal{A}_{D}, f_{i})$ in Eq.~\eqref{eq:eq3} {\footnotesize{\color{DarkBlue}{\tcp{mask matrix}}}}
        
        $\equrf, \equnrf \leftarrow (\mathcal{M}_{i}, f_{i})$ in Eq.~\eqref{eq:eq4} {\footnotesize{\color{DarkBlue}{\tcp{decoupling}}}}

        $\mathcal{L}_{DFD}(\equrf, \equnrf, y_{i}, \widehat{y_{i}}, \mathbf{m})_{i \in b} \leftarrow$ Eq.~\eqref{eq:eq5}

        {\footnotesize{\color{DarkBlue}{\tcc{DFC Module (Sec.\ref{sec42})}}}}

        $\equcorf \leftarrow (\mathcal{A}_{C}, \mathcal{M}_{i}, \equnrf)$ in Eq.~\eqref{eq:eq6} {\footnotesize{\color{DarkBlue}{\tcp{calibrate}}}}

        $\mathcal{L}_{DFC}(\equcorf, y_{i}, \mathbf{m})_{i \in b} \leftarrow$ Eq.~\eqref{eq:eq7}

        $\mathcal{L}_{CE}(\ell_{i}, y_{i})_{i \in b} \leftarrow$ Eq.~\eqref{eq:eq8}

        $\mathcal{L} = \mathcal{L}_{CE} + \frac{1}{|L|} \sum_{j=1}^{|L|} (\lambda_{1} \cdot \mathcal{L}_{DFD}^{L_{j}} + \lambda_{2} \cdot \mathcal{L}_{DFC}^{L_{j}})$

        $w_{k} \leftarrow w_{k} - \eta \nabla \mathcal{L}(w_{k};b)$ {\footnotesize{\color{DarkBlue}{\tcp{update model}}}}
    }
}

$w_{k} \leftarrow w_{k} - \{\mathcal{A}_{D}, \mathcal{A}_{C}, \mathbf{m}\}$ {\footnotesize{\color{DarkBlue}{\tcp{keep locally}}}}

return $w_{k}$ to the server

\end{algorithm}

\section{Experiments}
\subsection{Experimental Setup} \label{sec51}
\paragraph{Datasets.} We conduct our experiments on the following three popular multi-domain datasets:
\begin{itemize}
    \item Digits \cite{peng2019moment} includes four domains as MNIST (M), USPS (U), SVHN (SV), and SYN (SY), with $10$ categories.
    \item Office-Caltech \cite{gong2012geodesic} consists of four domains as Caltech (C), Amazon (A), Webcam (W), and DSLR (D), each is formed of $10$ overlapping categories.
    \item PACS \cite{li2017deeper} consists of four domains as Photo (P), Art-Painting (AP), Cartoon (Ct), Sketch (Sk). Each domain contains $7$ categories.
\end{itemize}

\paragraph{Domain-Skewed FL Scenarios.} 
Following~\cite{chen2024fair}, we initialize $20$, $10$, and $10$ clients for the Digits, Office-Caltech, and PACS, respectively. 
The allocation of domains to these clients is randomly generated and then fixed: Digits (M: $3$, U: $6$, SV: $6$, SY: $5$), Office-Caltech (C: $3$, A: $2$, W: $2$, D: $3$), PACS (P: $2$, AP: $3$, Ct: $2$, Sk: $3$). 
For each client, the local data is randomly selected from the allocated domains with $1\%$ (Digits), $20\%$ (Office-Caltech), and $30\%$ (PACS).

\paragraph{Model Architecture.} 
To facilitate a fair comparison, we use ResNet-10 \cite{he2016deep} to conduct experiments, with the dimension $d=512$. All compared methods use the same model.

\paragraph{Compared Methods.} We compare \my with four types of state-of-the-art (SOTA) baselines: 
1) standard FedAvg \cite{mcmahan2017communication}, FedProx \cite{li2020fedprox}; 2) contrast-based MOON \cite{li2021model}, FedRCL \cite{seo2024relaxed}; 3) prototype-based FPL \cite{huang2023rethinking}, FedTGP \cite{zhang2024fedtgp}, FedSA \cite{zhou2025fedsa}; and 4) consensus-based FedHEAL \cite{chen2024fair}, FDSE \cite{wang2025federated}. 
Among these methods, FPL \cite{huang2023rethinking}, FedHEAL \cite{chen2024fair}, and FDSE \cite{wang2025federated} are designed for the domain-skewed FL.

\paragraph{Implement Details.} 
Unless otherwise mentioned, we run $R=100$ global communication rounds, and the number of clients $K$ is set to $20$, $10$, and $10$ for Digits, Office-Caltech, and PACS datasets, respectively. 
All methods have little or no accuracy gain with more rounds $R$. We follow \cite{li2020fedprox,li2021model} and use the SGD optimizer to update models with a learning rate $\eta$ of $0.01$, a momentum of $0.9$, a batch size of $64$, and a weight decay of $10^{-5}$, with local epochs as $E=10$. The dimension $d$ of the embedding is $512$. 
We empirically set the $\sigma$ in Eq.~\eqref{eq:eq3} as $0.1$, the $\tau$ in Eq.~\eqref{eq:eq5} as $0.06$, $\lambda_{1}=0.8$ and $\lambda_{2}=1.0$ in Eq.~\eqref{eq:eq9}, $\alpha = 1.0$ and $\beta = 0.4$ in Eq.~\eqref{eq:eq11}. 
We implement the decoupler $\mathcal{A}_{D}$ in Sec.~\ref{sec41} and corrector $\mathcal{A}_{C}$ in Sec.~\ref{sec42} as a two-layer CNN architecture with batch normalization and ReLU activation, the input and output are both $\in \mathbb{R}^{C \times H \times W}$. The $\mathbf{m}$ in Eq.~\eqref{eq:eq5} and Eq.~\eqref{eq:eq7} is a single-layer MLP architecture to predict the logits as $\mathbb{R}^{d} \rightarrow \mathbb{R}^{\mathbb{C}}$. 
By default, we insert $\mathcal{A}_{D}$ and $\mathcal{A}_{C}$ after the last backbone layer of the feature extractor of the local model.

\paragraph{Evaluation Metrics.} 
Following \cite{li2020fedprox,li2019fair}, we use top-$1$ accuracy in each domain, and average (AVG) and standard deviation (STD) of accuracy across domains as evaluation metrics. All results reported are the average of three independent runs with different random seeds.

\begin{table*}[t!]\small
\caption{\textbf{Comparison results} on Office-Caltech and PACS. Best results are in \textcolor{DeepRed}{\textbf{red bold}}, with second best in \textcolor{blue}{\textbf{blue bold}}.}
\label{tab:tab1}
\centering
\setlength\tabcolsep{3pt} 
\renewcommand\arraystretch{0.88} 
\setlength{\arrayrulewidth}{0.2mm} 
\resizebox{\linewidth}{!}{
\begin{tabular}{l||cccc|cc I cccc|cc}
\hline\thickhline
\rowcolor{mygray} 
& \multicolumn{6}{c I}{\textbf{Office-Caltech}} & \multicolumn{6}{c}{\textbf{PACS}} \\
\cline{2-13}
\rowcolor{mygray} 
\multirow{-1.75}{*}{\centering \textbf{Methods}} & Caltech & Amazon & Webcam & DSLR & AVG $\pmb{\uparrow}$ & STD $\pmb{\downarrow}$ 
& Photo & Art-Paint & Cartoon & Sketch & AVG $\pmb{\uparrow}$ & STD $\pmb{\downarrow}$ \\
\hline\hline
FedAvg~\pub{AISTATS'17} 
& 59.82 & 65.26 & 51.72 & 46.67 & 55.86 & 8.27 
& 60.98 & 52.49 & 74.63 & 77.46 & 66.39 & 11.74 \\
FedProx~\pub{MLSys'20} 
& 60.18 & 66.11 & 54.96 & 47.30 & 57.13 & 7.98 
& 61.88 & 57.40 & 76.34 & 80.14 & 68.94 & 11.00 \\
MOON~\pub{CVPR'21} 
& 57.14 & 63.16 & 44.48 & 40.87 & 51.41 & 10.49 
& 59.72 & 50.29 & 70.35 & 70.22 & 62.64 & 9.63 \\
FPL~\pub{CVPR'23} 
& 60.05 & 65.53 & 52.90 & 64.45 & 60.73 & 5.75 
& 66.67 & 58.38 & 79.75 & 77.59 & 70.59 & 9.95 \\
FedTGP~\pub{AAAI'24} 
& 61.28 & 63.84 & 53.37 & 63.68 & 60.54 & 4.92 
& 60.68 & 56.06 & 75.27 & 78.03 & 67.51 & 10.78 \\
FedRCL~\pub{CVPR'24} 
& 60.27 & 65.89 & 48.72 & 50.14 & 56.24 & 8.64 
& 61.37 & 55.17 & 70.78 & 76.57 & 65.97 & 9.54 \\
FedHEAL~\pub{CVPR'24} 
& 61.02 & 67.47 & 55.36 & 62.99 & 61.71 & 4.95 
& 70.39 & 65.82 & 80.57 & 76.61 & \textcolor{blue}{\textbf{73.34}} & \textcolor{blue}{\textbf{6.54}} \\
FedSA~\pub{AAAI'25} 
& 63.52 & 66.21 & 56.90 & 58.93 & 61.39 & \textcolor{blue}{\textbf{4.24}} 
& 68.55 & 60.42 & 78.32 & 75.50 & 70.69 & 7.98 \\
FDSE~\pub{CVPR'25} 
& 60.39 & 66.80 & 58.31 & 67.20 & \textcolor{blue}{\textbf{63.18}} & 4.50 
& 69.27 & 65.46 & 78.65 & 79.12 & 73.13 & 6.83 \\
\hline
\rowcolor{myyellow} 
\my (Ours) 
& 62.95 & 68.42 & 64.79 & 71.12 & \textcolor{DeepRed}{\textbf{66.82}} & \textcolor{DeepRed}{\textbf{3.65}} 
& 75.15 & 68.71 & 79.91 & 82.11 & \textcolor{DeepRed}{\textbf{76.47}} & \textcolor{DeepRed}{\textbf{5.83}} \\
\hline\thickhline
\end{tabular}}
\end{table*}

\begin{table}[t]\small
\caption{\textbf{Comparison results} on Digits, continuing Table \ref{tab:tab1}.}
\label{tab:tab2}
\centering
\setlength\tabcolsep{2.88pt} 
\renewcommand\arraystretch{0.96} 
\setlength{\arrayrulewidth}{0.2mm} 
\resizebox{\linewidth}{!}{
\begin{tabular}{l||cccc|cc}
\hline\thickhline
\rowcolor{mygray} 
& \multicolumn{6}{c}{\textbf{Digits}} \\
\cline{2-7}
\rowcolor{mygray} 
\multirow{-1.75}{*}{\centering \textbf{Methods}} 
& MNIST & USPS & SVHN & SYN & AVG $\pmb{\uparrow}$ & STD $\pmb{\downarrow}$ \\
\hline\hline
FedAvg 
& 96.04 & 89.84 & 88.04 & 51.05 & 81.24 & 20.42 \\
FedProx 
& 96.13 & 89.04 & 87.56 & 51.00 & 80.93 & 20.30 \\
MOON 
& 94.70 & 89.64 & 87.68 & 34.40 & 76.60 & 28.29 \\
FPL 
& 96.12 & 89.94 & 89.15 & 49.15 & 81.09 & 21.52 \\
FedTGP 
& 95.38 & 87.49 & 87.37 & 49.40 & 79.91 & 20.68 \\
FedRCL 
& 93.08 & 89.73 & 84.98 & 51.20 & 79.75 & 19.32 \\
FedHEAL 
& 96.39 & 87.00 & 89.41 & 56.95 & 82.43 & 17.45 \\
FedSA 
& 94.85 & 90.74 & 88.48 & 57.41 & 82.87 & 17.18 \\
FDSE 
& 95.17 & 90.34 & 90.98 & 60.13 & \textcolor{blue}{\textbf{84.15}} & \textcolor{blue}{\textbf{16.19}} \\
\hline
\rowcolor{myyellow} 
\my (Ours)
& 97.75 & 92.94 & 90.53 & 67.69 & \textcolor{DeepRed}{\textbf{87.23}} & \textcolor{DeepRed}{\textbf{13.36}} \\
\hline\thickhline
\end{tabular}}
\end{table}

\subsection{Main Results} \label{sec52}
\paragraph{Comparison to SOTA Methods.} 
Table~\ref{tab:tab1} and Table~\ref{tab:tab2} provide the comparison results of our \my and nine competing methods on Digits, Office-Caltech, and PACS. 
These results empirically indicate that \my consistently outperforms other methods, which confirms that our method acquires well-generalizable ability across different domains and promotes cross-domain fairness (\ie, with a smaller STD). 
Additionally, we observe that contrast-based methods (\eg, MOON, FedRCL) perform poorly, even significantly worse than FedAvg on PACS. 
This may be because forcing the local models to align with compromised global representations through a contrastive scheme is detrimental and further exacerbates the performance degradation. 
This underscores the importance of promoting local training away from client-specific domain biases. 
Unlike the elimination-based, domain-skewed FL method FDSE, \my liberates additional class-relevant clues by calibrating domain-specific biased features, enabling more correct decisions across different domains.

\begin{table}[t]\small
\caption{\textbf{Modularity} of \my on PACS, where we plug-in our proposed DFD and DFC into four strong baselines.}
\label{tab:tab3}
\centering
\setlength\tabcolsep{2.0pt} 
\renewcommand\arraystretch{1.16} 
\setlength{\arrayrulewidth}{0.2mm} 
\resizebox{\linewidth}{!}{
\begin{tabular}{l||cccc|c}
\hline\thickhline
\rowcolor{mygray} 
\textbf{Methods} 
& P & AP & Ct & Sk & AVG $\pmb{\uparrow}$ \\
\hline\hline
FedAvg 
& 74.86 \myred{13.88} 
& 65.90 \myred{13.41} 
& 79.70 \myred{5.07} 
& 80.86 \myred{3.40} 
& 75.33 \myred{8.94} \\
FPL 
& 76.85 \myred{10.18} 
& 64.46 \myred{6.08} 
& 80.02 \myred{0.27} 
& 80.75 \myred{3.16} 
& 75.52 \myred{4.93} \\
FedHEAL 
& 75.56 \myred{5.17} 
& 65.95 \myred{0.13} 
& 81.27 \myred{0.70} 
& 77.48 \myred{0.87} 
& 75.06 \myred{1.72} \\
FDSE 
& 72.25 \myred{2.98} 
& 65.99 \myred{0.53} 
& 80.56 \myred{1.91} 
& 80.36 \myred{1.24} 
& 74.79 \myred{1.66} \\
\hline\thickhline
\end{tabular}}
\end{table}

\begin{figure}[t]
    \centering
    \setlength{\abovecaptionskip}{0.1cm}
    \setlength{\belowcaptionskip}{-0.2cm}
    \includegraphics[width=0.975\linewidth]{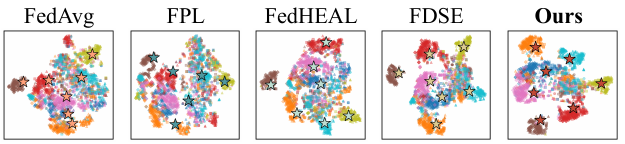}
    \caption{\textbf{T-SNE visualization} on PACS. 
    Each color means one class, each shape means one domain, stars are semantic centers.}
    \label{fig:fig4}
\end{figure}

\paragraph{Improving Performance of Existing FL Methods.} 
One key advantage of \my is its modularity. 
As shown in Table \ref{tab:tab3}, we report the results of baselines combined with our DFD and DFC modules, detailing both the resulting accuracy and the performance gain over the original methods. The results indicate that our two modules consistently enhance performance in multiple domains. FedAvg yields the largest benefit and achieves $8.94\%$ average accuracy gain over its original version.

\paragraph{T-SNE Visualization.} 
Fig.~\ref{fig:fig4} depicts the t-SNE \cite{tsne2008} visualization of the representation space for different FL methods on PACS. 
Compared to these four strong baselines, 1) \my enhances the inter-class separability, \ie, greater distance between samples of different colors compared to other methods; and 2) \my reduces the intra-class distance among samples of the same color, resulting in tighter clustering. 
This further indicates \my's ability to establish consistent semantic representations and decision boundaries across diverse domains.

\begin{table}[t]\small
\caption{\textbf{Efficiency comparison} per round ($10$ clients) on PACS.}
\label{tab:tab4}
\centering
\setlength\tabcolsep{5.6pt} 
\renewcommand\arraystretch{1.15} 
\setlength{\arrayrulewidth}{0.2mm} 
    \resizebox{\linewidth}{!}{
        \begin{tabular}{l||cccc}
        \hline\thickhline
        \rowcolor{mygray} 
        \textbf{Methods} & $\text{Upload}_{\times 10^{7}}$ & $\text{Comm.}$ 
        & $\text{FLOPs}_{\times 10^{10}}$ & $\text{Time}_{train}$ \\
        \hline\hline
        FedAvg 
            & 5.43 & 193.72MB & 16.31 & 176.94s \\
        FPL 
            & 5.45 & 197.30MB & 19.45 & 215.61s \\
        FedHEAL 
            & 5.43 & 193.72MB & 16.70 & 186.92s \\
        \rowcolor{myyellow} \my 
            & 5.43 & 193.72MB & 16.46 & 180.67s \\
        \hline\thickhline
        \end{tabular}
    }
\end{table}

\begin{figure}[t]
    \centering
    \setlength{\abovecaptionskip}{0.12cm}
    \setlength{\belowcaptionskip}{-0.16cm}
    \begin{minipage}[t]{0.495\linewidth}
        \centering
        \includegraphics[width=1.0\linewidth]{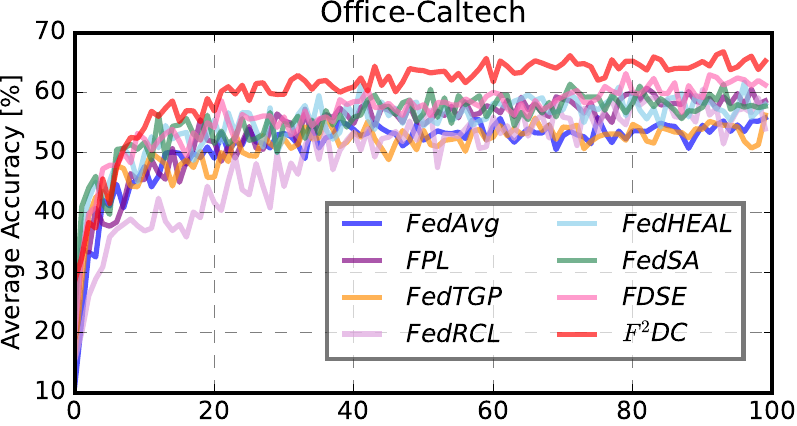}
    \end{minipage}
    \begin{minipage}[t]{0.495\linewidth}
        \centering
        \includegraphics[width=1.0\linewidth]{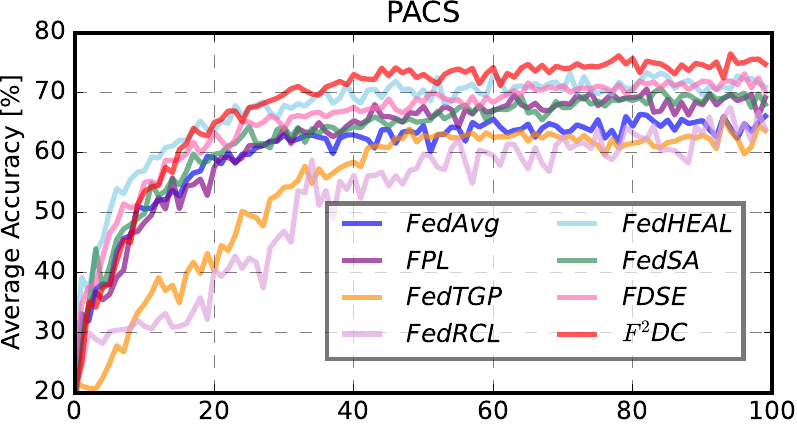}
    \end{minipage}
\caption{\textbf{Comparison of convergence in average accuracy} on Office-Caltech (\textbf{Left}) and PACS (\textbf{Right}). Zoom in for details.}
\label{fig:fig5}
\end{figure}

\paragraph{Efficiency Analysis.} 
Table~\ref{tab:tab4} provides the communication and computation cost on the PACS dataset with $10$ clients. Compared to FedAvg, \my incurs no additional communication cost. 
\my attains superior performance with a modest computational cost, where the additional training cost mainly stems from the decoupler and corrector.

\paragraph{Convergence.} 
In Fig.~\ref{fig:fig5}, we plot the average accuracy per round for various FL methods. 
It can be observed that our \my exhibits a faster convergence speed, and ultimately achieves higher performance as training progresses.

\subsection{Validation Analysis} \label{sec53}

\begin{figure}[t]
    \centering
    \setlength{\abovecaptionskip}{0.12cm}
    \setlength{\belowcaptionskip}{-0.26cm}
    \includegraphics[width=1.0\linewidth]{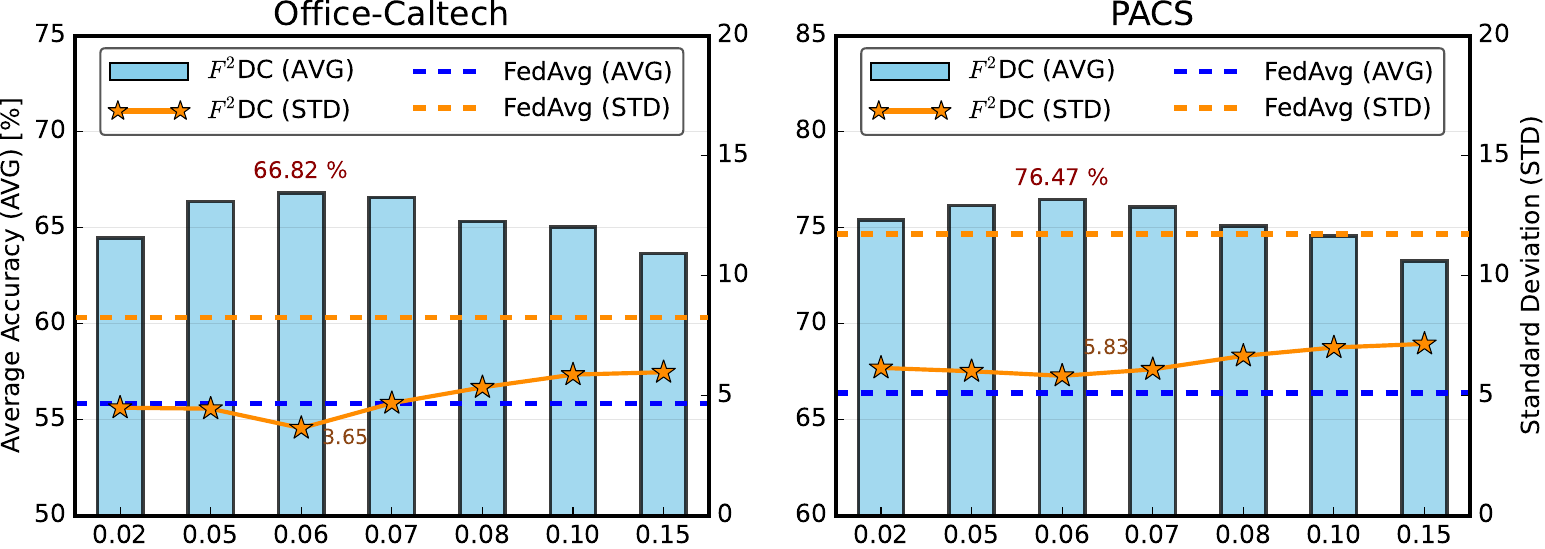}
    \caption{\textbf{Hyper-parameter study} with different $\tau$ of Eq.~\eqref{eq:eq5}.}
    \label{fig:fig6}
\end{figure}

\begin{table}[t]
\small
    \caption{\textbf{Hyper-parameter study} with $\sigma$ of Eq.~\eqref{eq:eq3} on PACS.}
    \label{tab:tab5}
    \centering
    \setlength\tabcolsep{4pt}
    \renewcommand\arraystretch{1.0}
    \setlength{\arrayrulewidth}{0.16mm}
    \begin{tabular}{c||ccccccc}
    \hline\thickhline
    \rowcolor{mygray}
    $\sigma$ & 0.01 & 0.05 & 0.1 & 0.3 & 0.5 & 1.0 & 5.0 \\
    \hline\hline
    AVG $\pmb{\uparrow}$ & 74.81 & 75.19 & \textbf{76.47} & 75.34 & 74.40 & 74.15 & 73.86 \\
    STD $\pmb{\downarrow}$ & 6.48 & 6.32 & \textbf{5.83} & 6.35 & 6.71 & 7.05 & 7.23 \\
    \hline\thickhline
    \end{tabular}
\end{table}

\begin{figure}[t]
    \centering
    \setlength{\abovecaptionskip}{0.cm}
    \setlength{\belowcaptionskip}{-0.26cm}
    \includegraphics[width=1.0\linewidth]{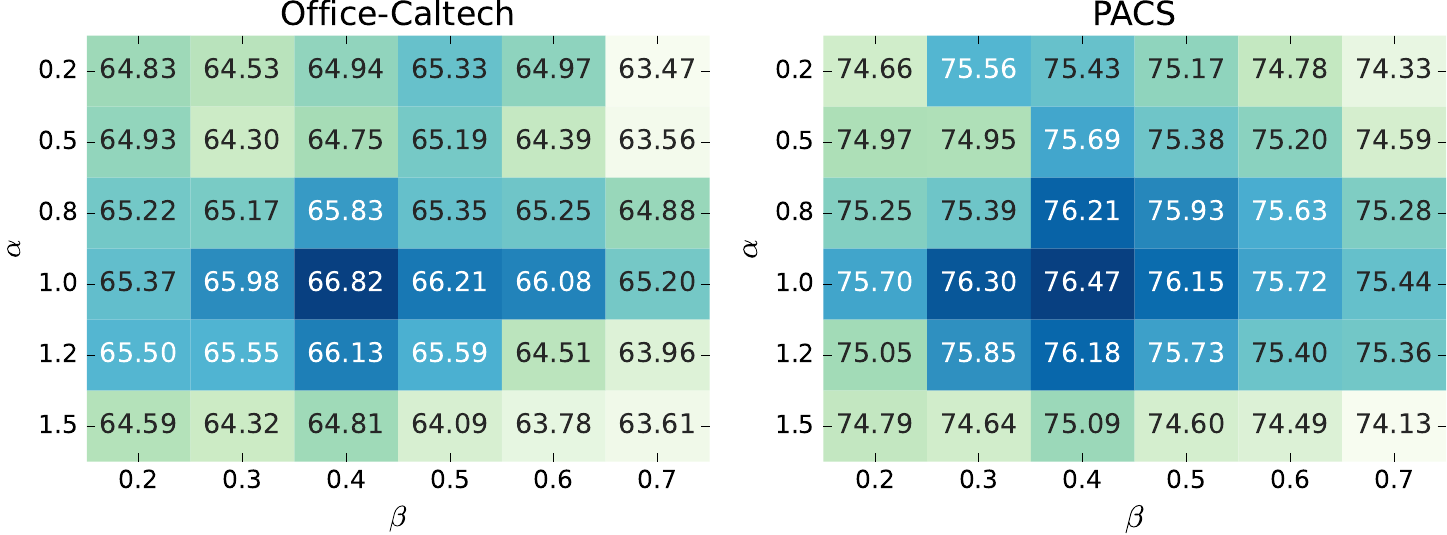}
    \caption{\textbf{Hyper-parameter study} with $\alpha$ and $\beta$ of Eq.~\eqref{eq:eq11}.}
    \label{fig:fig7}
\end{figure}

\begin{figure}[t]
    \centering
    \setlength{\abovecaptionskip}{0.cm}
    \setlength{\belowcaptionskip}{-0.26cm}
    \includegraphics[width=0.976\linewidth]{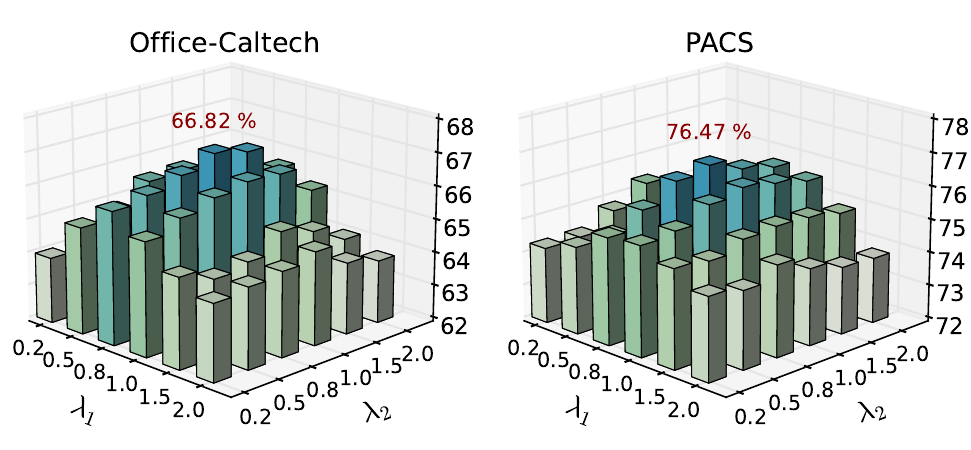}
    \caption{\textbf{Hyper-parameter study} with $\lambda_{1}$ and $\lambda_{2}$ of Eq.~\eqref{eq:eq9}.}
    \label{fig:fig8}
\end{figure}

\paragraph{Hyper-parameter Study.} 
We first examine the impact of hyper-parameter $\tau$ (Eq.~\eqref{eq:eq5}) in Fig.~\ref{fig:fig6}, which reveals that a smaller $\tau$ benefits feature separation more than higher ones. We observe that \my is not sensitive to $\tau \in [0.05, 0.07]$ and the optimal performance is reached when $\tau = 0.06$.

We then report the effect of hyper-parameter $\sigma$ (Eq.~\eqref{eq:eq3}) that controls the discreteness of the mask matrix in Table~\ref{tab:tab5}. 
For the low $\sigma$ values, mask matrix $\mathcal{M}_{i}$ of Eq.~\eqref{eq:eq3} becomes more discrete; for high values, $\mathcal{M}_{i}$ becomes more uniform. 
The value of $\sigma$ that is too small ($\rightarrow$ binary sampling) or too large ($\rightarrow$ uniform sampling) hinders training.

We investigate $\alpha$ and $\beta$ of Eq.~\eqref{eq:eq11} for a wide range, and show the accuracy in Fig.~\ref{fig:fig7}. We observe that \my brings performance improvement by incorporating domain discrepancy. 
Generally, $\alpha \in [0.8, 1.2]$ and $\beta \in [0.3, 0.5]$ is a suitable choice for domain-aware aggregation, leading to stable and competitive performance in \my. 
Besides, we present the accuracy for diverse $\lambda_{1}$ and $\lambda_{2}$ of Eq.~\eqref{eq:eq9} in Fig.~\ref{fig:fig8}. 
The results show that \my maintains consistent performance across varying $\lambda_{1}$ and $\lambda_{2}$ values, but extremely large $\lambda_{1}$ or $\lambda_{2}$ lead to optimization instability. 
The accuracy amelioration becomes marginal when $\alpha=1.0, \beta=0.4$ in Fig.~\ref{fig:fig7} and $\lambda_{1}=0.8, \lambda_{2}=1.0$ in Fig.~\ref{fig:fig8}. 

\begin{table}[t]
\caption{\textbf{Ablation study} of key modules in our \my on Office-Caltech (\textbf{Top}) and PACS (\textbf{Bottom}). Please see details in Sec.~\ref{sec53}.}
\label{tab:tab6}
\centering
\setlength\tabcolsep{1.66pt} 
\renewcommand\arraystretch{0.96} 
\setlength{\arrayrulewidth}{0.16mm} 
\resizebox{\linewidth}{!}{
\begin{tabular}{ccc||cccc I cc}
\hline\thickhline
\rowcolor{mygray} 
$\mathcal{L}_{DFD}$ & $\mathcal{L}_{DFC}$ & \texttt{DaA} 
& C & A & W & D & AVG $\pmb{\uparrow}$ & STD $\pmb{\downarrow}$ \\
\hline\hline
& & 
& 59.82 & 65.26 & 51.72 & 46.67 & 55.86 & 8.27 \\
& & \ding{51} 
& 60.25 & 65.79 & 54.39 & 50.33 & 57.69 & 6.76 \\
\ding{51} & & \ding{51} 
& 66.68 & 67.53 & 56.55 & 60.02 & 62.70 & 5.30 \\
& \ding{51} & \ding{51} 
& 63.71 & 68.21 & 59.07 & 67.62 & 64.65 & 4.22 \\
\ding{51} & \ding{51} & 
& 60.84 & 67.79 & 62.10 & 70.45 & 65.29 & 4.57 \\
\rowcolor{myyellow} 
\ding{51} & \ding{51} & \ding{51} 
& 62.95 & 68.42 & 64.79 & 71.12 & \textbf{66.82} & \textbf{3.65} \\
\hline\thickhline
\rowcolor{mygray} 
$\mathcal{L}_{DFD}$ & $\mathcal{L}_{DFC}$ & \texttt{DaA} 
& P & AP & Ct & Sk & AVG $\pmb{\uparrow}$ & STD $\pmb{\downarrow}$ \\
\hline\hline
& & 
& 60.98 & 52.49 & 74.63 & 77.46 & 66.39 & 11.74 \\
& & \ding{51}
& 61.28 & 57.64 & 75.71 & 79.09 & 68.43 & 10.15 \\
\ding{51} & & \ding{51}
& 72.95 & 61.52 & 79.06 & 80.13 & 73.41 & 8.35 \\
& \ding{51} & \ding{51}
& 72.55 & 65.44 & 77.61 & 78.95 & 73.64 & 6.12 \\
\ding{51} & \ding{51} & 
& 74.86 & 65.90 & 79.70 & 80.86 & 75.33 & 6.80 \\
\rowcolor{myyellow} 
\ding{51} & \ding{51} & \ding{51} 
& 75.15 & 68.71 & 79.91 & 82.11 & \textbf{76.47} & \textbf{5.83} \\
\hline\thickhline
\end{tabular}}
\end{table}

\begin{table}[t]
\caption{\textbf{Ablation study} of \my's various features on PACS.}
\label{tab:tab7}
\centering
\setlength\tabcolsep{5pt} 
\renewcommand\arraystretch{1.0} 
\setlength{\arrayrulewidth}{0.2mm} 
\resizebox{\linewidth}{!}{
\begin{tabular}{c||cccc I cc}
\hline\thickhline
\rowcolor{mygray} 
Protocols & P & AP & Ct & Sk & AVG $\pmb{\uparrow}$ & STD $\pmb{\downarrow}$ \\
\hline\hline
$f^{\scalebox{0.66}{$\bm{+}$}}$ & 74.21 & 67.52 & 78.35 & 80.47 & 75.13 & 5.90 \\
$f^{\scalebox{0.66}{$\bm{-}$}}$ & 59.45 & 50.09 & 52.56 & 69.37 & 57.87 & 8.63 \\
$f^{\scalebox{0.88}{$\star$}}$ & 74.83 & 64.05 & 75.80 & 79.29 & 73.49 & 6.71 \\
\rowcolor{myyellow} 
$\widetilde{f}$ & 75.15 & 68.71 & 79.91 & 82.11 & \textbf{76.47} & \textbf{5.83} \\
\hline\thickhline
\end{tabular}}
\end{table}

\paragraph{Ablation Study.} 
We conduct an ablation study in Table~\ref{tab:tab6} (first row refers to FedAvg \cite{mcmahan2017communication}). 
We observe that each component contributes positively to the overall performance, and their combination yields the optimal performance. This supports our motivation to decouple local features via DFD and then calibrate the domain-biased features via DFC.

In Table~\ref{tab:tab7}, we then report the results of using various features obtained by \my on PACS. Results show that $f^{\scalebox{0.66}{$\bm{+}$}}$ is significantly better than $f^{\scalebox{0.66}{$\bm{-}$}}$, indicating our decoupler effectively disentangles local features. Compared to $f^{\scalebox{0.66}{$\bm{-}$}}$, $f^{\scalebox{0.88}{$\star$}}$ yields consistent improvements, suggesting that $f^{\scalebox{0.88}{$\star$}}$ captures additional class-relevant clues for cross-domain decisions.


\section{Conclusion}
In this paper, we propose Federated Feature Decoupling and Calibration (\my) for domain-skewed FL. To achieve this, \my disentangles local features and corrects domain-biased features to capture additional useful clues, while performing domain-aware aggregation to promote consensus among clients. 
Extensive experimental results demonstrate the effectiveness and modularity of our proposed \my.

\section*{Acknowledgment}
This work is partially supported by the Australian Research Council Linkage Project LP210300009 and LP230100083, the Singapore Ministry of Education (MOE) Academic Research Fund (AcRF) Tier 1 Grant (24-SIS-SMU-008), A*STAR under its MTC YIRG Grant (M24N8c0103), and the Lee Kong Chian Fellowship (T050273).

{
    \small
    \bibliographystyle{ieeenat_fullname}
    \bibliography{main}

@String(PAMI = {IEEE Trans. Pattern Anal. Mach. Intell.})

@String(CVPR= {IEEE Conf. Comput. Vis. Pattern Recog.})

@String(ICCV= {Int. Conf. Comput. Vis.})

@String(ECCV= {Eur. Conf. Comput. Vis.})

@String(NIPS= {Adv. Neural Inform. Process. Syst.})

@String(ICME = {Int. Conf. Multimedia and Expo})

@String(ICLR = {Int. Conf. Learn. Represent.})

@String(IJCAI = {IJCAI})

@String(AAAI = {AAAI})

@String(PAMI  = {IEEE TPAMI})

@String(CVPR  = {CVPR})

@String(ICCV  = {ICCV})

@String(ECCV  = {ECCV})

@String(NIPS  = {NeurIPS})

@String(ICME  =	{ICME})

@String(ICLR  = {ICLR})

@String(ICML  = {ICML})

@String(ICDE = {ICDE})

@String(AISTATS = {AISTATS})

@String(MLSys = {MLSys})

@String(KDD = {ACM SIGKDD})

@inproceedings{mcmahan2017communication,
  title={Communication-efficient learning of deep networks from decentralized data},
  author={McMahan, Brendan and Moore, Eider and Ramage, Daniel and Hampson, Seth and y Arcas, Blaise Aguera},
  booktitle=AISTATS,
  pages={1273--1282},
  year={2017}
}

@article{li2020federated,
  title={Federated learning: Challenges, methods, and future directions},
  author={Li, Tian and Sahu, Anit Kumar and Talwalkar, Ameet and Smith, Virginia},
  journal={IEEE Signal Processing Magazine},
  volume={37},
  number={3},
  pages={50--60},
  year={2020}
}

@article{kairouz2021advances,
  title={Advances and open problems in federated learning},
  author={Kairouz, Peter and McMahan, H Brendan and Avent, Brendan and Bellet, Aur{\'e}lien and others},
  journal={Foundations and Trends in Machine Learning},
  volume={14},
  number={1},
  pages={1--210},
  year={2021}
}

@article{fu2025federated,
  title={Federated domain-independent prototype learning with alignments of representation and parameter spaces for feature shift},
  author={Fu, Lele and Huang, Sheng and Lai, Yanyi and Zhang, Chuanfu and Dai, Hong-Ning and Zheng, Zibin and Chen, Chuan},
  journal={IEEE Transactions on Mobile Computing},
  year={2025},
  volume={24},
  number={9},
  pages={9004--9019}
}

@article{de2005tutorial,
  title={A tutorial on the cross-entropy method},
  author={Pieter-Tjerk, De Boer and Kroese, Dirk P and Mannor, Shie and Rubinstein, Reuven},
  journal={Annals of Operations Research},
  volume={134},
  number={1},
  pages={19--67},
  year={2005}
}

@article{zhu2021federated,
  title={Federated learning on non-IID data: A survey},
  author={Zhu, Hangyu and Xu, Jinjin and Liu, Shiqing and Jin, Yaochu},
  journal={Elsevier Neurocomputing},
  volume={465},
  pages={371--390},
  year={2021}
}

@inproceedings{li2022federated,
  title={Federated learning on non-iid data silos: An experimental study},
  author={Li, Qinbin and Diao, Yiqun and Chen, Quan and He, Bingsheng},
  booktitle=ICDE,
  pages={965--978},
  year={2022}
}

@inproceedings{karimireddy2020scaffold,
  title={Scaffold: Stochastic controlled averaging for federated learning},
  author={Karimireddy, Sai Praneeth and Kale, Satyen and Mohri, Mehryar and Reddi, Sashank and Stich, Sebastian and Suresh, Ananda Theertha},
  booktitle=ICML,
  pages={5132--5143},
  year={2020}
}

@inproceedings{zhang2023dyted,
  title={Dyted: Disentangled representation learning for discrete-time dynamic graph},
  author={Zhang, Kaike and Cao, Qi and Fang, Gaolin and Xu, Bingbing and Zou, Hongjian and Shen, Huawei and Cheng, Xueqi},
  booktitle=KDD,
  pages={3309--3320},
  year={2023}
}

@article{bercea2022federated,
  title={Federated disentangled representation learning for unsupervised brain anomaly detection},
  author={Bercea, Cosmin I and Wiestler, Benedikt and Rueckert, Daniel and Albarqouni, Shadi},
  journal={Nature Machine Intelligence},
  volume={4},
  number={8},
  pages={685--695},
  year={2022}
}

@inproceedings{bai2024diprompt,
  title={Diprompt: Disentangled prompt tuning for multiple latent domain generalization in federated learning},
  author={Bai, Sikai and Zhang, Jie and Guo, Song and Li, Shuaicheng and Guo, Jingcai and Hou, Jun and Han, Tao and Lu, Xiaocheng},
  booktitle=CVPR,
  pages={27284--27293},
  year={2024}
}

@article{uddin2022federated,
  title={Federated learning via disentangled information bottleneck},
  author={Uddin, Md Palash and Xiang, Yong and Lu, Xuequan and Yearwood, John and Gao, Longxiang},
  journal={IEEE Transactions on Services Computing},
  volume={16},
  number={3},
  pages={1874--1889},
  year={2022}
}

@inproceedings{wang2025fedskc,
  title={FedSKC: Federated Learning with Non-IID Data via Structural Knowledge Collaboration},
  author={Wang, Huan and Li, Haoran and Chen, Huaming and Yan, Jun and Wang, Lijuan and Shi, Jiahua and Chen, Shiping and Shen, Jun},
  booktitle={IEEE International Conference on Web Services},
  pages={702--712},
  year={2025}
}

@inproceedings{zhou2025fedsa,
  title={Fedsa: A unified representation learning via semantic anchors for prototype-based federated learning},
  author={Zhou, Yanbing and Qu, Xiangmou and You, Chenlong and Zhou, Jiyang and Tang, Jingyue and Zheng, Xin and Cai, Chunmao and Wu, Yingbo},
  booktitle=AAAI,
  pages={23009--23017},
  year={2025}
}

@inproceedings{zhang2024fedtgp,
  title={Fedtgp: Trainable global prototypes with adaptive-margin-enhanced contrastive learning for data and model heterogeneity in federated learning},
  author={Zhang, Jianqing and Liu, Yang and Hua, Yang and Cao, Jian},
  booktitle=AAAI,
  pages={16768--16776},
  year={2024}
}

@inproceedings{he2016deep,
  title={Deep residual learning for image recognition},
  author={He, Kaiming and Zhang, Xiangyu and Ren, Shaoqing and Sun, Jian},
  booktitle=CVPR,
  pages={770--778},
  year={2016}
}

@inproceedings{li2020fedprox,
  title={Federated optimization in heterogeneous networks},
  author={Li, Tian and Sahu, Anit Kumar and Zaheer, Manzil and Sanjabi, Maziar and Talwalkar, Ameet and Smith, Virginia},
  booktitle=MLSys,
  pages={429--450},
  year={2020}
}

@inproceedings{wang2025feddifrc,
  title={FedDifRC: Unlocking the Potential of Text-to-Image Diffusion Models in Heterogeneous Federated Learning},
  author={Wang, Huan and Li, Haoran and Chen, Huaming and Yan, Jun and Shi, Jiahua and Shen, Jun},
  booktitle=ICCV,
  pages={3726--3736},
  year={2025}
}

@inproceedings{wang2025fedsc,
  title={FedSC: Federated Learning with Semantic-Aware Collaboration},
  author={Wang, Huan and Li, Haoran and Chen, Huaming and Yan, Jun and Shi, Jiahua and Shen, Jun},
  booktitle=KDD,
  pages={2938--2949},
  year={2025}
}

@inproceedings{huang2025fedbg,
  title={FedBG: Proactively mitigating bias in cross-domain graph federated learning using background data},
  author={Huang, Sheng and Fu, Lele and Liao, Tianchi and Deng, Bowen and Zhang, Chuanfu and Chen, Chuan},
  booktitle=IJCAI,
  pages={5408--5416},
  year={2025}
}

@inproceedings{xu2025federated,
  title={Federated learning with sample-level client drift mitigation},
  author={Xu, Haoran and Li, Jiaze and Wu, Wanyi and Ren, Hao},
  booktitle=AAAI,
  pages={21752--21760},
  year={2025}
}

@inproceedings{Li2020On,
  title={On the Convergence of FedAvg on Non-IID Data},
  author={Li, Xiang and Huang, Kaixuan and Yang, Wenhao and Wang, Shusen and Zhang, Zhihua},
  booktitle=ICLR,
  year={2020}
}

@inproceedings{zhang2023fedala,
  title={Fedala: Adaptive local aggregation for personalized federated learning},
  author={Zhang, Jianqing and Hua, Yang and Wang, Hao and Song, Tao and Xue, Zhengui and Ma, Ruhui and Guan, Haibing},
  booktitle=AAAI,
  pages={11237--11244},
  year={2023}
}

@inproceedings{ye2023feddisco,
  title={Feddisco: Federated learning with discrepancy-aware collaboration},
  author={Ye, Rui and Xu, Mingkai and Wang, Jianyu and Xu, Chenxin and Chen, Siheng and Wang, Yanfeng},
  booktitle=ICML,
  pages={39879--39902},
  year={2023}
}

@inproceedings{wang2021understanding,
  title={Understanding the behaviour of contrastive loss},
  author={Wang, Feng and Liu, Huaping},
  booktitle=CVPR,
  pages={2495--2504},
  year={2021}
}

@inproceedings{li2021model,
  title={Model-contrastive federated learning},
  author={Li, Qinbin and He, Bingsheng and Song, Dawn},
  booktitle=CVPR,
  pages={10713--10722},
  year={2021}
}

@inproceedings{li2017deeper,
  title={Deeper, broader and artier domain generalization},
  author={Li, Da and Yang, Yongxin and Song, Yi-Zhe and Hospedales, Timothy M},
  booktitle=ICCV,
  pages={5542--5550},
  year={2017}
}

@inproceedings{selvaraju2017grad,
  title={Grad-cam: Visual explanations from deep networks via gradient-based localization},
  author={Selvaraju, Ramprasaath R and Cogswell, Michael and Das, Abhishek and Vedantam, Ramakrishna and Parikh, Devi and Batra, Dhruv},
  booktitle=ICCV,
  pages={618--626},
  year={2017}
}

@inproceedings{wang2025federated,
  title={Federated Learning with Domain Shift Eraser},
  author={Wang, Zheng and Wang, Zihui and Fan, Xiaoliang and Wang, Cheng},
  booktitle=CVPR,
  pages={4978--4987},
  year={2025}
}

@inproceedings{gao2022feddc,
  title={Feddc: Federated learning with non-iid data via local drift decoupling and correction},
  author={Gao, Liang and Fu, Huazhu and Li, Li and Chen, Yingwen and Xu, Ming and Xu, Cheng-Zhong},
  booktitle=CVPR,
  pages={10112--10121},
  year={2022}
}

@article{huang2024federated,
  title={Federated learning for generalization, robustness, fairness: A survey and benchmark},
  author={Huang, Wenke and Ye, Mang and Shi, Zekun and Wan, Guancheng and Li, He and Du, Bo and Yang, Qiang},
  journal=PAMI,
  volume={46},
  number={12},
  pages={9387--9406},
  year={2024}
}

@inproceedings{huang2022learn,
  title={Learn from others and be yourself in heterogeneous federated learning},
  author={Huang, Wenke and Ye, Mang and Du, Bo},
  booktitle=CVPR,
  pages={10143--10153},
  year={2022}
}

@inproceedings{yan2023personalization,
  title={Personalization disentanglement for federated learning},
  author={Yan, Peng and Long, Guodong},
  booktitle=ICME,
  pages={318--323},
  year={2023}
}

@inproceedings{chen2024disentanglement,
  title={On disentanglement of asymmetrical knowledge transfer for modality-task agnostic federated learning},
  author={Chen, Jiayi and Zhang, Aidong},
  booktitle=AAAI,
  volume={38},
  number={10},
  pages={11311--11319},
  year={2024}
}

@inproceedings{cai2020rethinking,
  title={Rethinking differentiable search for mixed-precision neural networks},
  author={Cai, Zhaowei and Vasconcelos, Nuno},
  booktitle=CVPR,
  pages={2349--2358},
  year={2020}
}

@inproceedings{louizos2018learning,
  title={Learning Sparse Neural Networks through $L_0$ Regularization},
  author={Louizos, Christos and Welling, Max and Kingma, Diederik P},
  booktitle=ICLR,
  year={2018},
}

@inproceedings{wang2023fedcda,
  title={FedCDA: Federated learning with cross-rounds divergence-aware aggregation},
  author={Wang, Haozhao and Xu, Haoran and Li, Yichen and Xu, Yuan and others},
  booktitle=ICLR,
  year={2024}
}

@inproceedings{huang2023rethinking,
  title={Rethinking federated learning with domain shift: A prototype view},
  author={Huang, Wenke and Ye, Mang and Shi, Zekun and Li, He and Du, Bo},
  booktitle=CVPR,
  pages={16312--16322},
  year={2023}
}

@inproceedings{sun2021partialfed,
  title={Partialfed: Cross-domain personalized federated learning via partial initialization},
  author={Sun, Benyuan and Huo, Hongxing and Yang, Yi and Bai, Bo},
  booktitle=NIPS,
  pages={23309--23320},
  year={2021}
}

@inproceedings{wang2024aggregation,
  title={An aggregation-free federated learning for tackling data heterogeneity},
  author={Wang, Yuan and Fu, Huazhu and Kanagavelu, Renuga and Wei, Qingsong and others},
  booktitle=CVPR,
  pages={26233--26242},
  year={2024}
}

@inproceedings{chen2024fair,
  title={Fair federated learning under domain skew with local consistency and domain diversity},
  author={Chen, Yuhang and Huang, Wenke and Ye, Mang},
  booktitle=CVPR,
  pages={12077--12086},
  year={2024}
}

@inproceedings{jiang2022harmofl,
  title={Harmofl: Harmonizing local and global drifts in federated learning on heterogeneous medical images},
  author={Jiang, Meirui and Wang, Zirui and Dou, Qi},
  booktitle=AAAI,
  pages={1087--1095},
  year={2022}
}

@article{wang2024disentangled,
  title={Disentangled representation learning},
  author={Wang, Xin and Chen, Hong and Tang, Si'ao and Wu, Zihao and Zhu, Wenwu},
  journal=PAMI,
  volume={46},
  number={12},
  pages={9677--9696},
  year={2024}
}

@inproceedings{zhang2019gait,
  title={Gait recognition via disentangled representation learning},
  author={Zhang, Ziyuan and Tran, Luan and Yin, Xi and Atoum, Yousef and Liu, Xiaoming and Wan, Jian and Wang, Nanxin},
  booktitle=CVPR,
  pages={4710--4719},
  year={2019}
}

@inproceedings{fumero2023leveraging,
  title={Leveraging sparse and shared feature activations for disentangled representation learning},
  author={Fumero, Marco and Wenzel, Florian and Zancato, Luca and Achille, Alessandro and Rodol{\`a}, Emanuele and Soatto, Stefano and Sch{\"o}lkopf, Bernhard and Locatello, Francesco},
  booktitle=NIPS,
  pages={27682--27698},
  year={2023}
}

@inproceedings{ma2019learning,
  title={Learning disentangled representations for recommendation},
  author={Ma, Jianxin and Zhou, Chang and Cui, Peng and Yang, Hongxia and Zhu, Wenwu},
  booktitle=NIPS,
  pages={5711--5722},
  year={2019}
}

@inproceedings{mo2023disentangled,
  title={Disentangled multiplex graph representation learning},
  author={Mo, Yujie and Lei, Yajie and Shen, Jialie and Shi, Xiaoshuang and Shen, Heng Tao and Zhu, Xiaofeng},
  booktitle=ICML,
  pages={24983--25005},
  year={2023}
}

@inproceedings{li2020gait,
  title={Gait recognition via semi-supervised disentangled representation learning to identity and covariate features},
  author={Li, Xiang and Makihara, Yasushi and Xu, Chi and Yagi, Yasushi and Ren, Mingwu},
  booktitle=CVPR,
  pages={13309--13319},
  year={2020}
}

@inproceedings{t2020personalized,
  title={Personalized federated learning with moreau envelopes},
  author={T Dinh, Canh and Tran, Nguyen and Nguyen, Josh},
  booktitle=NIPS,
  pages={21394--21405},
  year={2020}
}

@inproceedings{li2025personalized,
  title={Personalized federated collaborative filtering: A variational autoencoder approach},
  author={Li, Zhiwei and Long, Guodong and Zhou, Tianyi and Jiang, Jing and Zhang, Chengqi},
  booktitle=AAAI,
  pages={18602--18610},
  year={2025}
}

@inproceedings{li2024personalized,
  title={Personalized federated domain-incremental learning based on adaptive knowledge matching},
  author={Li, Yichen and Xu, Wenchao and Wang, Haozhao and Qi, Yining and Guo, Jingcai and Li, Ruixuan},
  booktitle=ECCV,
  pages={127--144},
  year={2024}
}

@inproceedings{zhang2024fedgmkd,
  title={FedGMKD: An efficient prototype federated learning framework through knowledge distillation and discrepancy-aware aggregation},
  author={Zhang, Jianqiao and Shan, Caifeng and Han, Jungong},
  booktitle=NIPS,
  pages={118326--118356},
  year={2024}
}

@inproceedings{fu2025beyond,
  title={Beyond federated prototype learning: Learnable semantic anchors with hyperspherical contrast for domain-skewed data},
  author={Fu, Lele and Huang, Sheng and Lai, Yanyi and Liao, Tianchi and Zhang, Chuanfu and Chen, Chuan},
  booktitle=AAAI,
  pages={16648--16656},
  year={2025}
}

@inproceedings{peng2019moment,
  title={Moment matching for multi-source domain adaptation},
  author={Peng, Xingchao and Bai, Qinxun and Xia, Xide and Huang, Zijun and Saenko, Kate and Wang, Bo},
  booktitle=ICCV,
  pages={1406--1415},
  year={2019}
}

@inproceedings{gong2012geodesic,
  title={Geodesic flow kernel for unsupervised domain adaptation},
  author={Kong, Boqing and Shi, Yuan and Sha, Fei and Grauman, Kristen},
  booktitle=CVPR,
  pages={2066--2073},
  year={2012}
}

@inproceedings{seo2024relaxed,
  title={Relaxed contrastive learning for federated learning},
  author={Seo, Seonguk and Kim, Jinkyu and Kim, Geeho and Han, Bohyung},
  booktitle=CVPR,
  pages={12279--12288},
  year={2024}
}

@inproceedings{li2019fair,
  title={Fair resource allocation in federated learning},
  author={Li, Tian and Sanjabi, Maziar and Beirami, Ahmad and Smith, Virginia},
  booktitle=ICLR,
  year={2020}
}

@article{tsne2008,
  author  = {Laurens van der Maaten and Geoffrey Hinton},
  title   = {Visualizing Data using t-SNE},
  journal = {Journal of Machine Learning Research},
  year    = {2008},
  volume  = {9},
  number  = {86},
  pages   = {2579--2605}
}
}

\end{document}